\documentclass[journal]{IEEEtran}

\usepackage{cite}

\ifCLASSINFOpdf
  \usepackage[pdftex]{graphicx}
\else
  \usepackage[dvips]{graphicx}
\fi

\usepackage{amsmath}
\usepackage{amsfonts}
\usepackage{amssymb}
\usepackage{dsfont}
\DeclareMathOperator*{\argmax}{arg\,max}

\newcommand{\norm}[1]{\left\lVert#1\right\rVert}

\usepackage{enumitem}

\usepackage{multirow}

\usepackage{algorithmic}
\usepackage[linesnumbered,ruled]{algorithm2e}
\usepackage{xcolor}
\SetAlCapSkip{0.5em}
\newcommand{\st}[3][t]{\ensuremath{{#2}_{#3}^{#1}}}
\newcommand{\stvec}[2]{\ensuremath{\mathbf{#1}_{#2}}}
\newcommand{\traj}[2][1:N]{\ensuremath{\mathbf{\xi}_{#2}^{#1}}}
\newcommand{\comm}{\ensuremath{u_c}}
\newcommand{\lcomm}{\overline{u_c}}
\newcommand{\1}{\mathds{1}}
\newcommand{\prg}[1]{\noindent\textbf{#1. }} 

\begin{document}

\title{Efficient and Trustworthy Social Navigation Via Explicit and Implicit Robot-Human Communication}

\author{Yuhang~Che,
		Allison~M.~Okamura,~\IEEEmembership{Fellow,~IEEE,}
        and~Dorsa~Sadigh,~\IEEEmembership{Member,~IEEE} 
\thanks{This work was supported in part by the Human-centered AI seed grant program from the Stanford AI Lab, Stanford School of Medicine and Stanford Graduate School of Business.}
\thanks{Y. Che is with Waymo LLC, Mountain View, CA 94043. A. M. Okamura is with the Department of Mechanical Engineering, and D. Sadigh is with the Department of Computer Science, Stanford University, Stanford, CA 94305, USA. Email: {\tt\small yuhangc@waymo.com,  aokamura@stanford.edu, dorsa@cs.stanford.edu}.}%
}

\markboth{}%
{Authors \MakeLowercase{\textit{et al.}}: Efficient and Trustworthy Social Navigation Via Explicit and Implicit Robot-Human Communication}



\maketitle

\begin{abstract}
In this paper, we present a planning framework that uses a combination of implicit (robot motion) and explicit (visual/audio/haptic feedback) communication during mobile robot navigation. First, we developed a model that approximates both continuous movements and discrete behavior modes in human navigation, considering the effects of implicit and explicit communication on human decision making. The model approximates the human as an optimal agent, with a reward function obtained through inverse reinforcement learning. Second, a planner uses this model to generate communicative actions that maximize the robot's transparency and efficiency. We implemented the planner on a mobile robot, using a wearable haptic device for explicit communication. In a user study of an indoor human-robot pair orthogonal crossing situation, the robot was able to actively communicate its intent to users in order to avoid collisions and facilitate efficient trajectories. Results showed that the planner generated plans that were easier to understand, reduced users' effort, and increased users’ trust of the robot, compared to simply performing collision avoidance. The key contribution of this work is the integration and analysis of explicit communication (together with implicit communication) for social navigation.
\end{abstract}


\IEEEpeerreviewmaketitle

\section{Introduction} \label{sec: intro}
\IEEEPARstart{M}{obile} robots are entering human environments, with applications ranging from delivery and support in warehouses, to home and social services. In this work we focus on social navigation, in which the movements and decisions of robots and humans affect each other. Researchers have studied various methods for generating robot motions that can avoid collision and comply to social rules during navigation~\cite{Trautman2013, Ferrer2014, kollmitz2015, kretzschmar2016, kim2016, pfeiffer2016, chen2017}. These methods have shown great potential towards bringing intelligent robots into human environments. However, there is an inherent limitation: robot motion alone does not capture all aspects of social interaction. In scenarios like social navigation, there are many ways that humans interact with each other. For example, by establishing eye contact, two people often assume that they will both take actions to avoid each other (people often act more conservatively if they think others are not paying attention). In addition, there are scenarios where conflicts occur and require more involved communication. One common example is people choosing to move to the same side when trying to avoid each other, and then moving to the other side simultaneously again. In this case, a person usually has to stop and gesture the other person to go first. Social robots need to be able to handle such conflicts in a similar manner by incorporating different ways of communication.

\begin{figure}[t]
	\centering
    \includegraphics[width=0.45\textwidth]{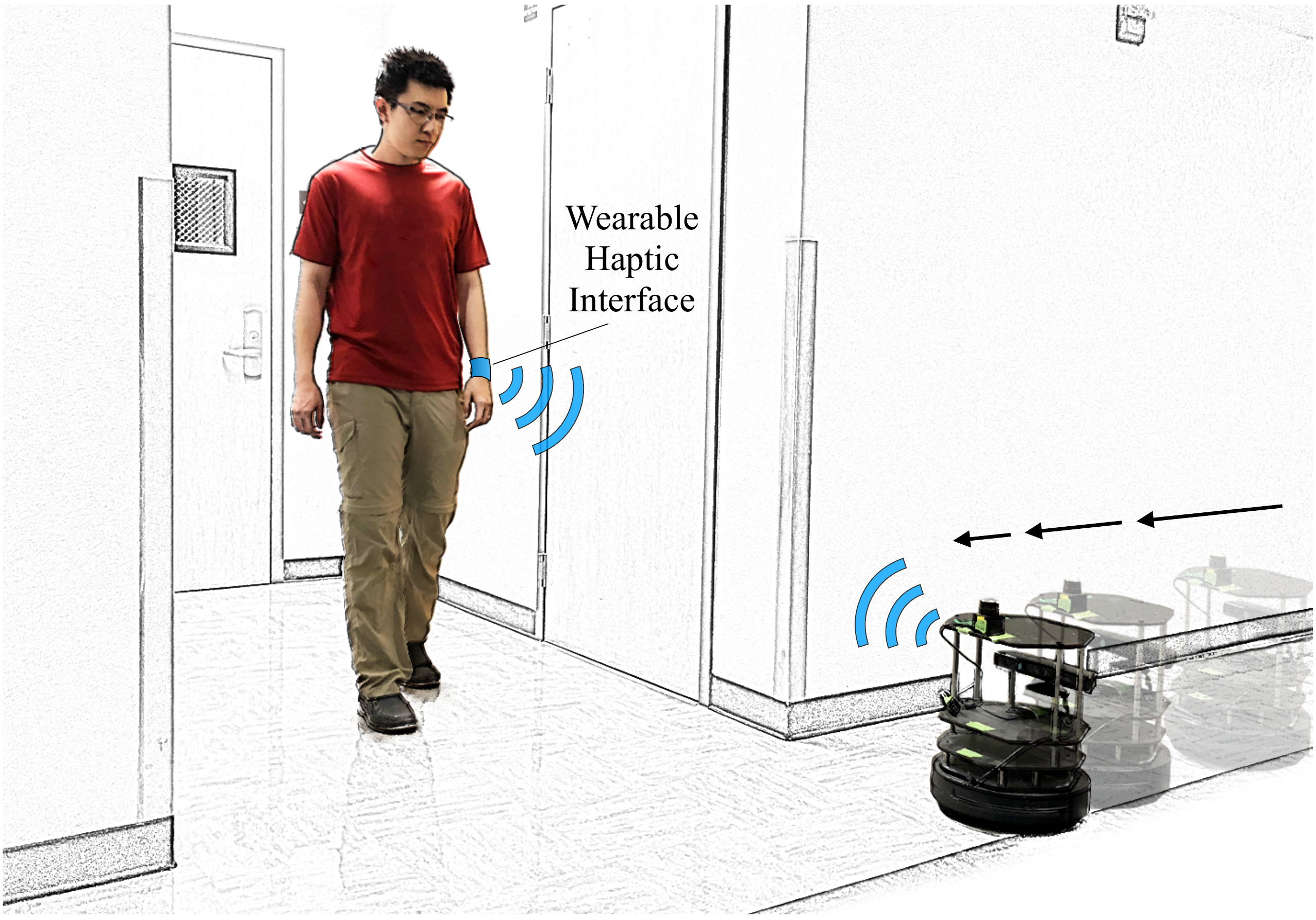}
    \caption{A social navigation scenario in which the robot communicates its intent (to yield to the human) to the human both implicitly and explicitly. Implicit communication is achieved via robot motion (slowing down and stopping), and explicit communication is achieved through a wearable haptic interface. }
    \label{fig: intro}
\end{figure}

The aforementioned limitation motivates us to explore richer interaction and communication schemes for robot social navigation. Specifically, we propose to consider both \textit{implicit communication} and \textit{explicit communication} (See Fig.~\ref{fig: intro}). In human-robot interaction literature, there are various definitions of implicit and explicit communication~\cite{peterson1959,breazeal2005,knepper2017}. Our definition of implicit communication closely follows that of~\cite{knepper2017}: the information from implicit communication need to be extracted by leveraging the context and common ground. Conversely, we define explicit communication as actions that convey information in a deliberate and non-ambiguous manner.

Examples of implicit communication include eye contact and body language. In social navigation, people often encode their intentions into their motion, which serves as implicit communication~\cite{mavrogiannis2016}. Similarly, a robot's motion carries information about its intent. The notion of motion as implicit communication is closely related to the idea of legibility~\cite{Dragan2013}, which measures how well a robot's motion (and behavior, more generally speaking) expresses its intent. In this work, we develop an algorithm that enables emergence of legible motions in social navigation.

Given our definition, explicit communication can take many forms, including spoken language, visual signals (symbols and colors) and audio signals (sirens), as long as the meanings of these signals are established beforehand. In this work, we focus on the use of haptic feedback because it is immediate and targeted. In addition, haptic feedback is advantageous when users' visual and auditory channels are already overloaded. However, the algorithms developed in this work are not limited to haptic communication and can be directly applied to other forms of explicit communication.

Both implicit and explicit communication during navigation will affect how people move. Thus, we propose that the ability to understand and predict navigation behaviors of humans would be beneficial to planning communicative actions. We developed a data-driven, probabilistic model to predict human behaviors, including both continuous movements and discrete behavior modes, given robot actions. Leveraging the human model, we develop an algorithm that plans for a robot's motion and communication through haptic feedback. Our approach relies on the assumption that users are cooperative -- they will not intentionally thwart the robot. In environments in which humans and robots work together, such as offices, homes, hospitals, and warehouses, this assumption is reasonable. We leave scenarios where humans act adversarially to future work. We also assume that users are equipped with wearable haptic interfaces, and the robot can send feedback via these interfaces (can be replaced with sound or visual cues if appropriate). In this work, we focus on the interaction between one robot and one human user.

The main contributions of this work are:
\begin{enumerate}
	\item A predictive model of human navigation behavior in response to both implicit and explicit robot communication. The model is described in Section IV, and the framework in which the model is applied is described in Section III.
    \item An interactive planning algorithm based on the human model that enables a robot to proactively communicate through either implicit or explicit communication, with the goal of efficient and transparent interaction (Section V).
    \item Implementation of the proposed algorithm on a physical mobile robot platform, and analysis and verification of the algorithm with user studies. We first evaluated and tuned the system in simulation (Section VI) and then performed experiments (Section VII).
\end{enumerate}

\section{Background} \label{sec: background}
\subsection{Social Navigation}
In traditional robot motion planning frameworks, humans are usually treated as obstacles that the robot should avoid~\cite{fox1997}. However, in social navigation scenarios, humans are different from most other obstacles -- they have a purpose, follow social rules, and react to other agents in the environment, including the robot~\cite{kruse2013}. Researchers have explored various methods for social-aware motion planning.

Understanding and modeling reactive behaviors of humans is necessary for planning socially acceptable actions. A popular framework to model human navigation in response to other agents is the Social Force Model (SFM)~\cite{helbing1995, zanlungo2011, farina2017}, which frames pedestrians as affected by interactive forces that drive them towards their goals and away from each other. SFM has been applied to human-robot interaction and social navigation scenarios~\cite{ratsamee2013, Ferrer2014, mehta2016, Che2018}. Some work in the field of multi-agent planning also consider the reactive nature of moving agents. For example, optimal reciprocal collision avoidance (ORCA)~\cite{van2011} assumes that each agent takes some responsibility of resolving pair-wise conflicts, which result in global collision free paths. Guy et al. extended ORCA to model human navigation by introducing additional parameters such as reaction time and personal space~\cite{guy2010}. These reaction-based methods are popular for crowd simulation because of the high computational efficiency. However, they tend to be less accurate in predicting individual trajectories, and do not capture the stochasticity of human behavior.

Researchers have applied probabilistic methods to modeling human behavior and planning for social navigation. The Interactive Gaussian Process model was proposed in~\cite{trautman2010unfreezing}; it takes into account the variability of human behavior and mutual interactions. Trautman et al. used this model to navigate a robot through a crowded cafeteria~\cite{Trautman2013}. Inverse Reinforcement Learning (IRL)~\cite{ng2000, levine2012, abbeel2005, ziebart2008maximum} is another framework that allows a probabilistic representation of human behavior. IRL was applied in~\cite{kuderer2012, kim2016, kretzschmar2016, pfeiffer2016} to learn pedestrian models and plan for socially compliant navigation. In general, these approaches aim to generate ``human-like'' actions, with the underlying assumption that the robot should behave similarly to a human in order to be socially acceptable. We also use IRL as part of our model to predict humans' responses. Our work differs from previous approaches in that we model jointly the effects of both explicit communication and robot motions on human navigation behavior.

Recently, deep neural networks also gained attention in robot navigation applications. Chen et al. incorporated social norms and multi-agent reactions into a deep reinforcement learning framework to enable robot navigation in a pedestrian-rich environment~\cite{chen2017}. Tai et al. applied a generative adversarial imitation learning strategy to generate socially compliant navigation from raw sensor data~\cite{tail2018}. Deep learning frameworks have the advantage that they often impose less assumptions on human behavior. However, it is unclear whether they generate motions that are natural and transparent. In addition, the amount of data required for training these neural networks can be large and expensive to obtain. Our approach is more data efficient and explainable by leveraging modeling techniques.

\subsection{Expressive Motion in Robotics}
Besides social navigation, researchers have investigated the problem of planning interactive and communicative motions in the field of computational human-robot interaction~\cite{thomaz2016}. Dragan et al. formalized the idea of legibility in~\cite{dragan2013legibility} and proposed methods to generate legible motions for robot manipulators~\cite{Dragan2013}. Legible motions are motions that express the robot's intent to observers, and can be different from predictable motions~\cite{dragan2013legibility}. Sadigh et al. modeled a human-robot system jointly as a dynamical system, and proposed methods to plan motions that actively influence human behaviors~\cite{sadigh2016planning} and further actively gather information of internal state of humans~\cite{sadigh2016information,sadigh2018planning}. Similar ideas were explored in human-robot collaboration scenarios. Bestick et al. developed a method to enable a robot to purposefully influence the choices of a person in handover tasks to help the person avoid suboptimal grasps~\cite{bestick2016}. Nikolaidis et al. studied human-robot mutual adaptation, and proposed a decision framework to allow the robot to guide participants towards a better way of completing the task~\cite{nikolaidis2017}. In our work, we consider robot motions that can inform or affect people as a form of implicit communication, as people often have to infer about the robot's intent with context.

\subsection{Non-Verbal Explicit Communication Methods}
In this work, we define actions that communicate information deliberately and unambiguously as explicit communication. The most common form of explicit communication is verbal communication. Non-verbal communication can be made explicit, such as pointing gestures~\cite{huang2013, lohse2014}, head nod~\cite{hashimoto2007} and visual displays~\cite{baraka2018}.

Haptic feedback has been used as an explicit communication mechanism in human-robot interaction. Scheggi et al. designed a vibrotactile haptic bracelet to guide the user along trajectories that are feasible for human leader-robot follower formation tasks~\cite{scheggi2012, scheggi2014}. The bracelet can display three distinct signals that inform human to slow down and turn toward left or right. Sieber et al. used haptic feedback to assist a user teleoperating a team of robots manipulating an object~\cite{sieber2015} by communicating the status (stable vs. in transient) of the robot team. Here, we use haptic feedback to explicitly convey the robot's discrete intent to the user in collision avoidance scenarios during navigation.

Non-verbal communication, such as gaze, are considered as implicit if it indirectly conveys information~\cite{mutlu2009, admoni2014, moon2014}. In this work, we use the motion of the mobile robot, which does not impose any constraint on the appearance of the robot and can naturally be incorporated in social navigation.

\section{A Planning Framework \\for Socially-Aware Robot Navigation}\label{sec: planning framework}
\begin{figure}[t]
	\centering
    \includegraphics[width=0.35\textwidth]{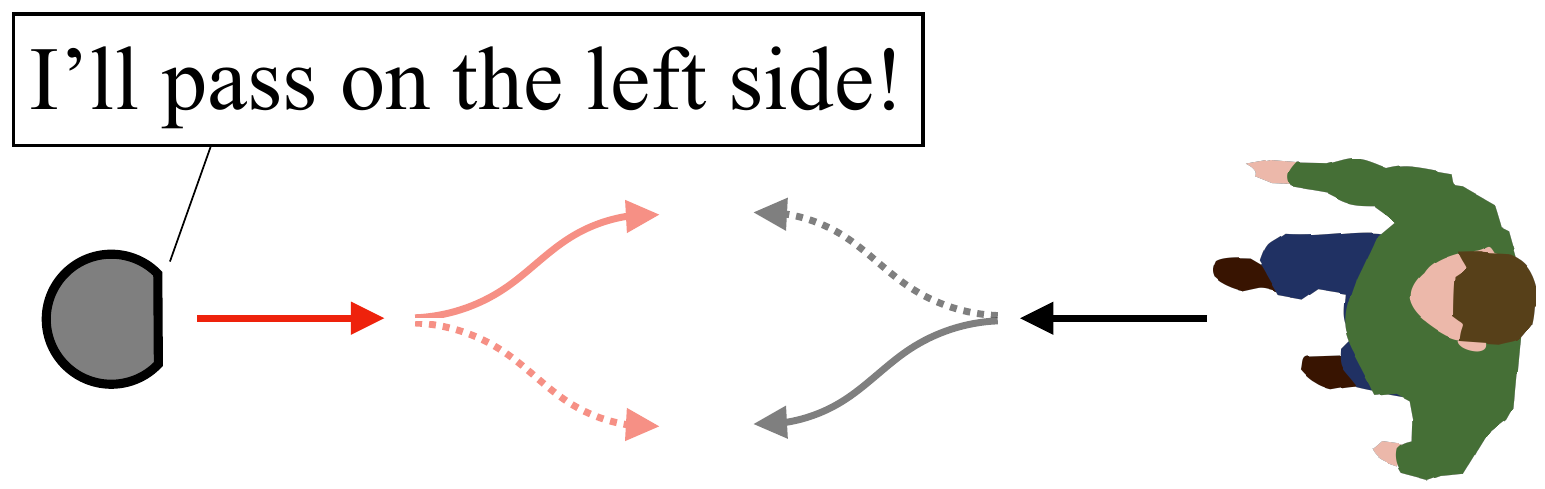}
    \caption{An example scenario where the robot needs to plan for appropriate movement and communication.}
    \label{fig: running example}
\end{figure}

Humans have an impressive ability to navigate seamlessly around other people. They walk still walk smoothly and efficiently when faced with a large group of people walking toward or around them in busy streets, concerts, cafeterias, conferences, or office spaces. However, robots lack the same \emph{elegance} while navigating around people. Our goal is to design mobile robots that navigate in a similar fashion to humans in spaces shared by humans and robots.
\emph{Our insight for social navigation is that mobile robots -- similar to humans -- need to pro-actively plan for interactions with people.}
In this section, we provide a high level outline of our interaction framework, and we will provide a more detailed description in Sections~\ref{sec: human model} and~\ref{sec: planning}.

In this work, we focus on a scenario in which one robot interacts with one human. 
To illustrate this, we use the running example illustrated in Fig.~\ref{fig: running example}: a robot and a human need to pass each other either on the left side or on the right side. To achieve this pass smoothly, the robot and human must understand each other's intent and coordinate their movements. The objective of our planning framework is to generate appropriate explicit communication and robot motions (which also serve as implicit communication) to facilitate human-robot interactions in such scenarios. The explicit communication consists of a finite number of discrete signals, for example, expressing a plan to pass on the left or the right side. The robot motions consist of continuous actions such as the robot's linear and angular velocities.

We model the joint human-robot system as a fully-observable interactive dynamical system in which the actions of the human and the robot can influence each other. In addition, we consider the stochasticity of actions of the human and the effect of communication in our planning framework.

Let $\st{s}{h}$ and $\st{s}{r}$ denote the physical state of the human and the robot at a certain time step $t$, respectively. The states are continuous, and include positions and velocities of each agent. The robot can apply continuous control inputs $\st{u}{r}$ that change its state through the following dynamics:
\begin{equation}
	\st{s}{r} = \mathcal{F}_r(\st[t-1]{s}{r}, \st{u}{r})
\end{equation}

Similarly, the state of the human is directly changed by her action $\st{u}{h}$:
\begin{equation}
	\st{s}{h} = \mathcal{F}_h(\st[t-1]{s}{h}, \st{u}{h})
\end{equation}

We let $\mathbf{u}_{r} = [\st[1]{u}{r}, ..., \st[N]{u}{r}]^\intercal$ and $\mathbf{u}_{h} = [\st[1]{u}{h}, ..., \st[N]{u}{h}]^\intercal$ denote the sequence of robot and human actions over a finite horizon $N$ ($[\cdot]^\intercal$ denotes a vector). $\traj{r} = [(\st[1]{s}{r}, \st[1]{u}{r}), ..., (\st[N]{s}{r}, \st[N]{u}{r})]^\intercal$ and $\traj{h} = [(\st[1]{s}{h}, \st[1]{u}{h}), ..., (\st[N]{s}{h}, \st[N]{u}{h})]^\intercal$ are then the state-action trajectories of the robot and the human over the horizon $N$, respectively. We use $\comm$ to refer to the explicit communication actions. Each of the actions $\st[i]{u}{r}$ and $\st[i]{u}{h}$ are vectors with appropriate dimensions. In this work, we focus on a one-dimensional communication action $\comm$, i.e., $\comm \in \mathcal{U}_c \cup \text{``none"}$.

We define an overall reward function for the robot $R_r(\traj{r}, \traj{h}, \comm)$. We let this reward function depend on both the robot's trajectory and the trajectory of the human. This reward function encodes properties such as efficiency (e.g., robot's velocity, distance to goal), safety (distance to human), and other metrics. 
In this work, we use an optimization-based planning algorithm, where the robot will optimize the expected sum of its reward function $R_r$.
Eq.~(\ref{eq: computing optimal u_r}) represents this expected reward, which is computed over predicted trajectories $\traj{h}$ of the human. We emphasize that the distribution of $\traj{h}$ is affected by the robot actions. In the next section, we present a hybrid model that predicts this distribution.


We will discuss our planning algorithm in Sec.~\ref{sec: planning}.
We will use Model Predictive Control (MPC)~\cite{morari1993} algorithm, where at every time step, the robot computes a finite horizon sequence of actions to maximize its expected reward:
\begin{equation}
\stvec{u}{r}^*, \comm^* = \argmax_{\stvec{u}{r}, \comm} \mathbb{E}_{\traj{h}} \left[ R_r(\traj{r}, \traj{h}, \comm) \right]
\label{eq: computing optimal u_r}
\end{equation}

\begin{figure*}[t]
	\centering
    \includegraphics[width=0.9\textwidth]{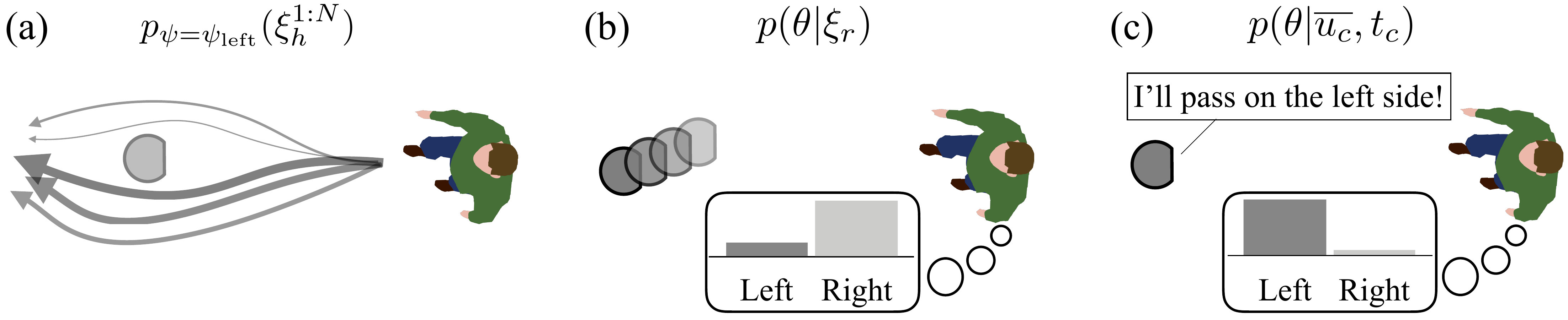}
    \caption{Overview of different components of the model. (a) We model the distribution of human trajectories given their behavior mode $\psi$. Each behavior mode is associated with an underlying intent. In this example, trajectories that pass the robot on the left have much higher probabilities (thicker lines), because the underlying intent is to pass on the left. (b) The robot's implicit communication (motion) affects the belief of the human over the robot's intent, and therefore affects the behavior mode of the human. (c) The robot's explicit communication also has strong effects on the belief and behavior mode of the human.}
    \label{fig: model overview}
\end{figure*}

In an MPC scheme, at each time step, the robot only executes the first control in the optimal sequence $\mathbf{u}_{r}^*$, and then replans at the next time step. The robot also plans on an explicit communication $\comm^* \in \mathcal{U}_c \cup \text{``none"}$. The robot decides to either communicate ($\comm^* \in \mathcal{U}_c$), or does not provide any communication ($\comm^* = \text{``none"}$). When computing the plan at each time step, we assume that explicit communication can only happen at the beginning of the planning horizon. This is because only the first set of actions in the planned sequence is executed, and a new plan will be generated in the next time step.

The key insight of this framework is that the robot is aware of the stochasticity of human behavior and the influences of communication on humans' actions. As a result, the robot will pro-actively communicate its intent to avoid undesirable situations. Take the example scenario in Fig.~\ref{fig: running example}: if the robot and the human choose to pass each other on different sides, a collision could happen. By expressing its intent explicitly, the robot minimizes the chance of a potential collision.

\section{Hybrid Model of Human Navigation} \label{sec: human model}
In this section, we present our model of human navigation, focusing on how the robot's implicit and explicit communication affect behaviors of the human.

To compute the optimal robot actions using Eq.~\eqref{eq: computing optimal u_r}, the robot is required to predict human actions over a finite time horizon. Since human actions are stochastic, the robot needs to predict the distribution of $\mathbf{u}_{h}$. We assume the human dynamics $\mathcal{F}_h$ are known. Therefore, the distribution of $\mathbf{u}_{h}$ induces an equivalent distribution over human trajectories $\traj{h}$ for a given $\mathbf{u}_{h}$ and initial state. Modeling this distribution exactly is quite challenging and may be computationally expensive. Our objective is to develop an approximate model that captures the interactive nature of human behavior, i.e., \emph{how the human will react to the robot's movements and explicit communication.}

Fig.~\ref{fig: model overview} illustrates the three most important components of our model. On a high level, we model human navigation behavior as a joint mixture of distribution over the trajectory, similar to~\cite{kretzschmar2016}:

\begin{equation}
p(\traj{h}) = \sum_{\psi} p_{\psi}(\traj{h}) p(\psi)
\label{eq: human model general}
\end{equation}

\noindent The first part of the mixture, $p_{\psi}(\traj{h})$, models the distribution of the continues state-actions given a specific behavior mode $\psi$. The second part, $p(\psi)$, models the distribution of the discrete behavior mode, $\psi$. The behavior mode reflects the underlying intent of the human, and is affected by both the robot's implicit communication (motion) $\xi_r$ and explicit communication $\lcomm$.

The idea of discrete behavior modes or classes of actions is sometimes associated with homotopy classes of trajectories~\cite{kretzschmar2016,mavrogiannis2018}. Here we do not enforce the trajectories given a specific value of $\psi$ to belong to one homotopy class. Instead, we associate each value of $\psi$ with an underlying reward function $R_h^\psi$ that the human optimizes. With an inverse reinforcement learning framework, our model assigns high probabilities to trajectories that comply with the behavior mode (one homotopy that corresponds to the underlying intent of the human), and low but non-zero probabilities to trajectories in other homotopy classes. This representation is more generalizable than homotopy, as different behavior modes are not necessarily different in homotopy in other scenarios. For example, tailgating versus keeping safe distances are two behavior modes in driving, and they cannot be differentiated by homotopy. However, they can be characterized by different cost functions that drivers may optimize.

For computational efficiency, we use relatively simple models for human and robot dynamics. The human state $\st{s}{h} = [x_h, y_h, \dot{x}_h, \dot{y}_h]^\intercal$ consists of positions and velocities in the 2D plane, and the robot state $\st{s}{r} = [x_r, y_r, \theta_r]^\intercal$ is simply its pose. We use a constant acceleration model for human dynamics. Thus, the human control input is $\st{u}{h} = [\ddot{x}_h, \ddot{y}_h]^\intercal$. For the robot dynamics, we use a differential drive kinematic model, with control input $\st{u}{r} = [v, \omega]^\intercal$, consisting of linear and angular velocities. 

The formulation in Eq.~(\ref{eq: human model general}) describes our modeling approach in its most general form. In our specific setup, the distribution $p(\traj{h})$ should be conditioned on the history of human and robot trajectories, history of explicit communication, and future robot actions over the prediction horizon. To numerically compute the distribution and use it for planning, we make two assumptions:\\

\prg{Assumption 1} The trajectory of the human $\traj{h}$ over the prediction horizon depends on the environment, $E$, (including the goal of the human) and the future robot trajectory $\traj{r}$:
\begin{equation}
    p_{\psi}(\traj{h} | \traj{r}, E)
    \label{eq: continuous model}
\end{equation}

\noindent Here actions of the human depend on the robot's actions. However, the robot also needs to predict actions of the human and plan accordingly. This inter-dependency leads to the \emph{infinite regress} problem, i.e., a sequence of reasonings about each others' actions that can never come to an end. To resolve this problem, we assume that the human has access to the future robot trajectory, and model the interaction between the human and the robot as a two-player Stackelberg game (leader-follower game). We argue that this assumption is reasonable: given relatively short planning horizon, humans are usually able to predict immediate actions of other agents. Since the focus of this work is the interaction between the robot and the human, the goals and environment are assumed to be known and fixed.\\


\prg{Assumption 2} We assume that the behavior mode is only affected by past information. With this, we can express the second term in Eq.~(\ref{eq: human model general}) as:
\begin{equation}
    p(\psi | \traj[-N:0]{r}, \traj[-N:0]{h}, u_c^{-N:0})
\end{equation}

\noindent where $\traj[-N:0]{r}, \traj[-N:0]{h}$ are the past trajectories of the robot and the human, and $u_c^{-N:0}$ is the past explicit communications over a horizon of length $N$. We further assume that only the most recent explicit communication actually affects the behavior mode:
\begin{equation}
    p(\psi | \traj[-N:0]{r}, \traj[-N:0]{h}, \lcomm, t_c)
    \label{eq: discrete model}
\end{equation}

Here $\lcomm$ represents the most recent explicit communication, and $t_c$ represents the time when $\lcomm$ is communicated. Note that at each time step, the robot can decide to not perform any explicit communication ($u_c = \text{``none"}$). The most recent communication refers to the most recent $\lcomm \neq \text{``none"} $, e.g., the last time haptic feedback was provided.

In the next two subsections, we discuss details of modeling humans' continuous navigation actions as in Eq.~(\ref{eq: continuous model}) and the humans' discrete behavior mode as in Eq.~(\ref{eq: discrete model}).

\subsection{Modeling Continuous Navigation Actions of a Human}

\prg{Inverse Reinforcement Learning}
We employ a data-driven approach to model humans' continuous navigation actions. Specifically, we use maximum entropy inverse reinforcement learning (MaxEnt IRL)~\cite{ziebart2008maximum, levine2012} to learn the distribution of human actions that matches a set of provided demonstrations in expectation. We assume the humans are approximately optimizing a reward function $R_{h}^\psi(\traj{h}, \traj{r}, E)$ for a given discrete class $\psi$. Under this model, the probability of the actions $\mathbf{u}_{h}$ is proportional to the exponential of the total reward of the trajectory ($Z$ is a normalization vector):
\begin{equation}
\begin{split}
p_\psi(\traj{h} | \traj{r}, E) &= \frac{1}{Z} \exp \left( R_{h}^\psi(\traj{h}, \traj{r}, E) \right) \\
&= \frac{1}{Z} \exp \left( \mathbf{w}_\psi \cdot \mathbf{f}(\traj{h}, \traj{r}, E) \right)
\end{split}
\label{eq: max ent}
\end{equation}

We parameterize the human reward function as a linear combination of features $\mathbf{f}$ that capture relevant properties of the navigation behavior. The features are weighted by $\mathbf{w}_\psi$, which can be learned by maximizing the overall likelihood of the provided demonstrations $\mathcal{D}$:
\begin{equation}
\argmax_{\mathbf{w}} \prod_{\mathbf{u}_{h} \in \mathcal{D}} p_\mathbf{w}(\traj{h} | \traj{r}, E)
\label{eq: irl}
\end{equation}

\noindent The weight vector $\mathbf{w}_\psi$ is a function of the behavior mode: $\psi$. Given a selected set of features such as distance to the other agents, heading, or velocity, we learn appropriate weights corresponding to the humans' reward functions for each discrete class $\psi$ from collected demonstrations. We will discuss the specific features used in this work in Sec.~\ref{sec: simulation}.

\subsection{Modeling Behavior Modes of a Human}
When modeling the distribution of behavior modes of a human, we consider the effect of both implicit communication (via robot movements) and explicit communication (via haptic feedback). We assume that the human will infer the robot's intent, and act cooperatively during the interaction. Mathematically, this suggests that the distribution of behavior modes of the human is related to her belief over the robot's intent, $\theta$. In our particular social navigation problem, we assume this belief is equal to probability of choosing a behavior mode:
\begin{equation}
p(\psi | \traj[-N:0]{r}, \traj[-N:0]{h}, \lcomm, t_c) = p(\theta | \traj[-N:0]{r}, \traj[-N:0]{h}, \lcomm, t_c)
\label{eq: belief model assumption}
\end{equation}
In other words, if the human believes that there is an 80\% chance that the robot will yield priority, she will behave in ``human priority'' mode with the same probability of 0.8. Non-linear relationships between $p(\psi)$ and $p(\theta)$ could also be used to model risk-averse (more willing to yield priority) and risk-seeking behaviors.

Using Bayes rule, we can transform Eq.~(\ref{eq: belief model assumption}) to:
\begin{equation}
\begin{split}
&p(\theta | \traj[-N:0]{r}, \traj[-N:0]{h}, \lcomm, t_c) 
\\
&\propto  p(\traj[-N:0]{r}, \lcomm, t_c | \theta, \traj[-N:0]{h}) \cdot p(\theta | \traj[-N:0]{h}) \\
 & = p(\traj[-N:0]{r} | \theta, \traj[-N:0]{h})\cdot p(\lcomm, t_c | \theta)\cdot p(\theta)
\end{split}
\label{eq: discrete model factorization}
\end{equation}

The second step in Eq.~(\ref{eq: discrete model factorization}) assumes conditional independence of the robot trajectory $\traj[-N:0]{r}$ and explicit communication $u_c$. With this factorization, we can separately model the effect of robot motion (implicit communication) and explicit communication on behavior modes of the human. The last term $p(\theta)$ is the prior on the robot's intent, which should be decided based on the application. In our implementation, there is an equal chance for the robot communicates different intents. So we choose a uniform prior.

The formulation casts the backward inference problem (from action to intent) into forward prediction problems (from intent to action). Next, we derive methods to compute each part of Eq.~\eqref{eq: discrete model factorization}.\\

\prg{Implicit Communication}
We assume that the human in general expects the robot to be rational and efficient according to some reward function. Applying the principle of maximum entropy, we model the human as expecting robot movements with probability:
\begin{equation}
    p(\traj[-N:0]{r} | \theta, \traj[-N:0]{h}) = \frac{\exp \left( R_r^\theta(\traj[-N:0]{r}, \traj[-N:0]{h}) \right)}{\int_{\mathbf{\xi}'_{r}} \exp \left( R_r^\theta(\mathbf{\xi}'_{r}, \traj[-N:0]{h}) \right) \mathrm{d} \mathbf{\xi}'_{r}}
    \label{eq: implicit model}
\end{equation}
where $R_r^\theta(\traj[-N:0]{r}, \traj[-N:0]{h})$ is a reward function that the human expects the robot to optimize, given its intent $\theta$. To compute the integration in the denominator of Eq.~\eqref{eq: implicit model}, we use the second-order Taylor series to approximate $R_r^\theta(\traj[-N:0]{r}, \traj[-N:0]{h})$.\\

\prg{Explicit Communication}
The belief of the human over the robot's intent is strongly affected by explicit communication, because the intent is directly conveyed. However, the effect of explicit communication should decay over time, as the robot's intent may change, and only short-term intents are communicated. Inspired by a model of human short-term verbal retention~\cite{peterson1959}, we propose:
\begin{equation}
    p(\lcomm, t_c | \theta) = \frac{1}{Z} \left\{
    \begin{array}{ll}
         A \exp \left( -\frac{t - t_c}{M} \right) + 1 & \lcomm = \theta \\
         1 & \lcomm \neq \theta
    \end{array}
    \right.
    \label{eq: explicit model}
\end{equation}

\noindent Here $A$ and $M$ are parameters that determine the characteristic of the distribution, and $Z$ is a normalization factor. 
The explicit communication initially reflects the true intent with very high probability. However, the inference strength decays over time, and eventually the communication becomes irrelevant to the robot's latest intent.

\begin{figure}[t]
	\centering
    \includegraphics[width=0.4\textwidth]{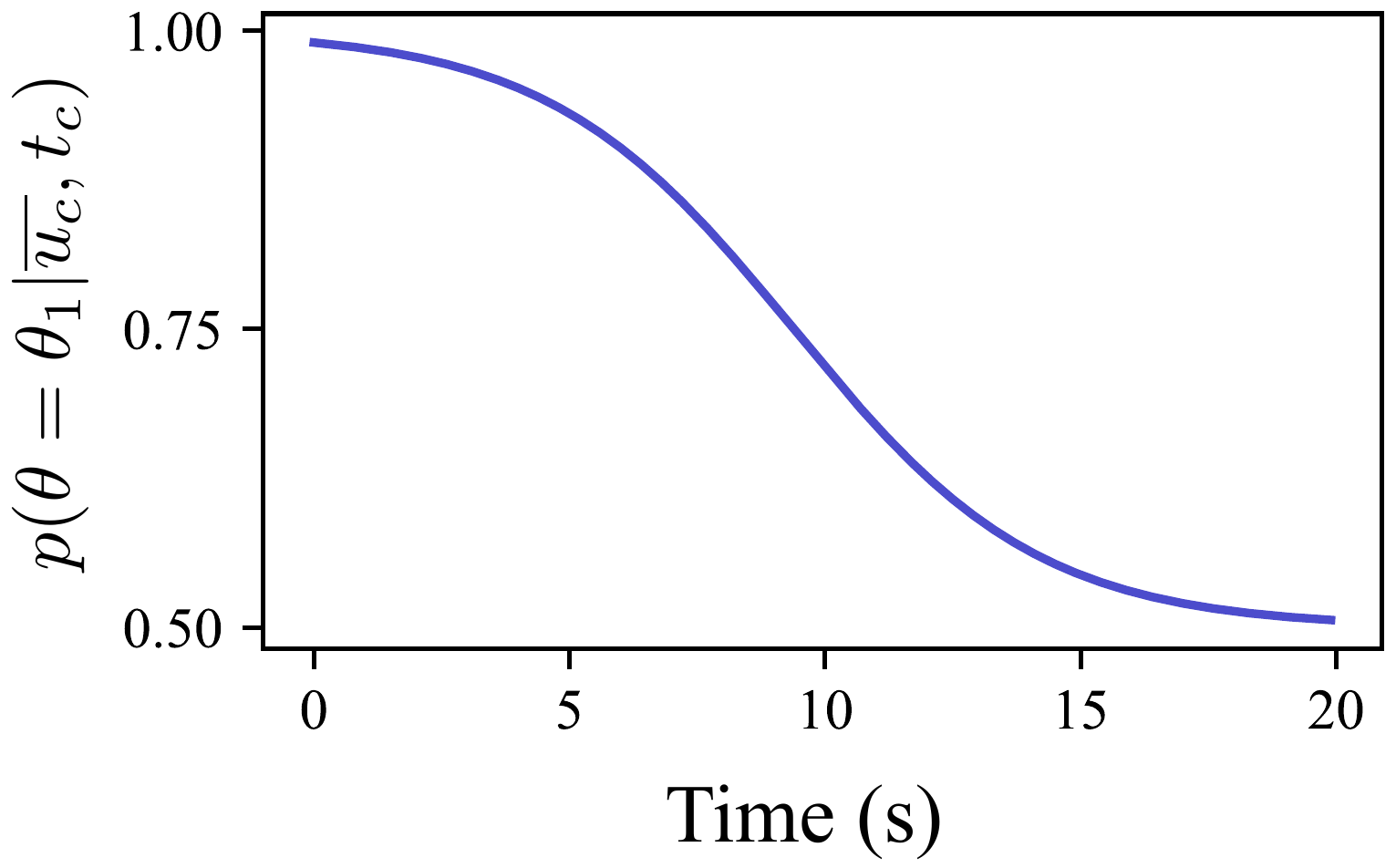}
    \caption{Inference on the robot's intent based on explicit communication. Assuming that $\lcomm = \theta_1$ happened at $t_c = 0$}
    \label{fig: explicit model demo}
\end{figure}

To clarify this model, let us consider a simpler version of Eq.~\eqref{eq: belief model assumption}: $P(\theta | \lcomm, t_c)$ -- to infer the robot's intent with only explicit communication. Assume that $\theta \in \{\theta_1, \theta_2\}$ is binary, $\lcomm= \theta_1$, and $t_c = 0$, and use Bayes rule again:
\begin{equation}
\begin{split}
    p(\theta = \theta_1 | \lcomm, t_c) &= \frac{p(\lcomm, t_c | \theta_1) p(\theta_1)}{p(\lcomm, t_c |\theta_1) p(\theta_1) + p(\lcomm, t_c | \theta_2) p(\theta_2)}\\
    &= \frac{A \exp \left(- (t - t_c) / {M} \right) + 1}{A \exp \left(- (t - t_c) / {M} \right) + 2}
\end{split}
\end{equation}
The change of $p(\theta = \theta_1 | \lcomm, t_c)$ over time is plotted in Fig.~\ref{fig: explicit model demo}. Initially, the robot's intent can be inferred with high probability given the explicit communication. As time passes, the inference strength decreases and eventually the belief over the robot's intent becomes uniform (0.5).

In the scenarios we consider, the human infers the robot's intent with both implicit and explicit communication. The combined effect can be computed with Eqs.~(\ref{eq: discrete model factorization})~--~(\ref{eq: explicit model}) given in this section.

\section{Planning for Communication} \label{sec: planning}
The general planning framework was described in Section~\ref{sec: planning framework}. In this section, we discuss the details of the algorithm, including the design of reward functions, derivation of the solution to the optimization, and an outline of our implementation.

\subsection{Robot Reward Function}
The overall reward $R_r$ that robot optimizes in Eq.~\eqref{eq: computing optimal u_r} consists of four parts that quantify \textit{robot efficiency}, \textit{human comfort}, \textit{safety}, and \textit{reward of explicit communication}:

\prg{Robot Efficiency} The robot should get as close as possible to its target with minimum effort:
    \begin{equation}
        R_{r,r} (\traj{r}) = \norm{\st[N]{x}{r} - x_{r}^{g}} +  \gamma_r \sum_{t=1}^{N} \norm{\st{u}{r}}^2,
    \end{equation}
    where $\gamma_r$ is a weight that balances effort and goal. This component determines how fast the robot approaches the goal. $\st[N]{x}{r}$ is the robot position at time step $N$, and $x_{r}^{g}$ is the goal position. $\norm{\cdot}$ denotes Euclidean norm.
    
\prg{Human Comfort} We want the human to spend less effort to achieve her goal:
    \begin{equation}
        R_{r,h} (\traj{h}) = \norm{\st[N]{x}{h} - x_{h}^{g}} + \gamma_h \sum_{t=1}^{N} \norm{\st{u}{r}}^2.
    \end{equation}
    Similar to $\gamma_r$, $\gamma_h$ balances between human reaching her goal and her effort. 
    
\prg{Safety} The robot should avoid collisions with the human:
    \begin{equation}
        R_{r,\text{safety}} (\traj{r}, \traj{h}) = \sum_{t=1}^{N} \exp \left( - \frac{\norm{\mathbf{x}_{h,t} - \mathbf{x}_{r,t}}^2}{D_r^2} \right).
    \end{equation}
    
\prg{Explicit Communication} Too much explicit communication will distract or annoy the human. Therefore, we set a constant reward for performing each explicit communication:
    \begin{equation}
        R_{r,EC} (u_c) = \gamma_{EC} \cdot \1(u_c \neq \text{``none"})
    \end{equation}
where $\1$ is the indicator function.


\noindent The components defined above are then combined linearly with weights $c_i$:
\begin{equation}
	R_r(\traj{r}, \traj{h}, \comm) = \sum_{i=0}^4 c_i \cdot R_{r,i}(\traj{r}, \traj{h}, \comm)
	\label{eq: robot reward function}
\end{equation}

\subsection{Human-Aware Planning}
To solve for the optimal robot actions using Eq.~\eqref{eq: computing optimal u_r}, we need to compute the expectation of the robot reward over the distribution of $\mathbf{u}_h$. In the last section, we derived a hybrid model for human behavior. To speed up computation, here we only consider the most likely trajectory given each possible human intent. With this, the expectation in Eq.~\eqref{eq: computing optimal u_r} can be expressed as:
\begin{equation}
    \mathbb{E}_{\traj{h}} \left[ R_r \right] = \sum_{\psi} p(\psi) \cdot R_r(\traj{r}, \traj[\psi]{h},\comm),
    \label{eq: expected reward}
\end{equation}
where $\traj[\psi]{h}$ is the most likely human response given $\psi$, and can be computed as: 
\begin{equation}
\traj[\psi]{h} = \argmax_{\traj{h}} \mathbf{w}_\psi \cdot \mathbf{f}(\traj{h}, \traj{r})
\label{eq: most likely actions}
\end{equation}
The distribution $p(\psi)$ is given by Eqs.~\eqref{eq: belief model assumption} and~\eqref{eq: discrete model factorization}.

We use gradient-based optimization to solve the optimal robot controls $\mathbf{u}_{r}^*$. This requires the gradient information of the expected reward in Eq.~\eqref{eq: expected reward}. Letting $\overline{R}_r = \mathbb{E}_{\traj{h}} \left[ R_r \right]$, we aim to find:
\begin{equation}
    \frac{\partial \overline{R}_r}{\partial \mathbf{u}_{r}} = \frac{\partial \overline{R}_r}{\partial \mathbf{u}_{r}} + \sum_{\psi} \frac{\partial \overline{R}_r}{\partial \mathbf{u}_{h}} \frac{\partial \mathbf{u}_{h}^{\psi}}{\partial \mathbf{u}_{r}},
\end{equation}
where $\mathbf{u}_{h}^{\psi}$ is the most likely human actions. As we have a symbolic representation of $R_r$, both $\frac{\partial \overline{R}_r}{\partial \mathbf{u}_{r}}$ and $\frac{\partial \overline{R}_r}{\partial \mathbf{u}_{h}}$ can be computed analytically. Following the implicit differentiation method discussed in~\cite{sadigh2016planning}, we compute the last unknown term using the following expression:
\begin{equation}
    \frac{\partial \mathbf{u}_{h}^{\psi}}{\partial \mathbf{u}_{r}} = \left[ \frac{\partial^2 R_h^{\psi}}{\partial \mathbf{u}_{h}^2} \right]^{-1} \left[ - \frac{\partial^2 R_h^{\psi}}{\partial \mathbf{u}_{h} \partial \mathbf{u}_{r}} \right]
\end{equation}
We have assumed a discrete set of communication actions $\mathcal{U}_c \cup \text{``none"}$.
To obtain the optimal explicit communication $u_c^*$, we enumerate all possible communication actions $u_c \in \mathcal{U}_c \cup \text{``none"}$ and solve the optimization for each $u_c$. This can be done in parallel because the optimizations do not depend on each other.

\begin{algorithm}[tb]
	\KwData{$\mathcal{U}_c$ - set of explicit communicative actions}
	\SetKwData{TimerI}{timer1}	\SetKwData{TimerII}{timer2}
	\SetKwFunction{SendHaptics}{SendHapticMessage}
    
    \SetKwFunction{GoalReached}{GoalReached}
    \SetKwFunction{GetTrackingInfo}{GetTrackingInfo}
    \SetKwFunction{UpdateModel}{UpdateModel}
    \SetKwFunction{PredictOneStep}{PredictOneStep}
    \SetKwFunction{GenerateInitGuess}{GenerateInitGuess}
    \SetKwFunction{ComputePlan}{ComputePlan}
	\SetKwFunction{Communicate}{Communicate}
    \SetKwFunction{ExecuteControl}{ExecuteControl}
    
    \SetKwProg{ForParallel}{for}{ do in parallel}{end}
	
    \Repeat {\GoalReached{$\st{x}{r}$, $x_{g, r}$, $\epsilon$}} {
    	\tcc{get current+predicted states}
    	$\st{s}{r}, \st{u}{r}, \st{s}{h} \gets$ \GetTrackingInfo{}\;
        $\st{\hat{s}}{r}, \st{\hat{s}}{h} \gets$ \PredictOneStep{$\st{s}{r}, \st{s}{h}$}\;
        
        \BlankLine
        \tcc{update the hybrid model}
        $p(\psi) \gets$ \UpdateModel{$p(\psi), \st{s}{r}, \st{u}{r}, \st{s}{h}, u'_c, t_c$}\;
        
        \BlankLine
        \tcc{find the optimal actions}
        $\mathbf{u}_{r}^{0}, \mathbf{u}_{h}^{0} \gets$ \GenerateInitGuess{$\st{\hat{s}}{r}, \st{\hat{s}}{h}, x_{g,r}, x_{g,h}$}\;
        
        \ForParallel {$u_c \in \mathcal{U}_c \cup \{null\}$ } {
        	$\mathbf{u}_r(u_c), \overline{R}_r(u_c) \gets$ \ComputePlan{$\mathbf{u}_{r}^{0}, \mathbf{u}_{h}^{0}, \st{\hat{s}}{r}, \st{\hat{s}}{h}, p(\psi), u_c$}\;
        }
        
        $u_c^* \gets \argmax_{u_c} \overline{R}_r(u_c)$ \;
        $\mathbf{u}_r^* = \mathbf{u}_r(u_c^*)$\;
        }
        
        \BlankLine
        \tcc{execute the actions}
        \If {$u_c^* \neq null$} {
        	\Communicate{$u_c^*$}\;
            $u'_c \gets u_c^*$\;
            $t_c \gets t$\;
        }
        
        \ExecuteControl{$\mathbf{u}_r^*$}\;
        
        \BlankLine
        $t \gets t + 1$\;

	\caption{Outline of the planning algorithm.}
	\label{alg: planning framework}
\end{algorithm}

The planning algorithm is outlined in Alg.~\ref{alg: planning framework}. At every time step, the algorithm first retrieves states of the robot and the human, and performs a one step prediction of robot and human states in order to compensate time spent for planning. Then it updates the belief over the behavior mode of the human $\psi$ with new observations using equations (\ref{eq: belief model assumption}) and (\ref{eq: discrete model factorization}). Before optimizing for robot actions, the algorithm needs to generate an initial guess for the initial state of the human. If there is a plan from the previous time step, the plan is used as the initial guess. When there is no previous plan (first time step, or first detection of human), we generate the initial guesses as follows: First, we compute the robot actions ignoring the human. We use a feedback control based policy to steer the robot towards its goal~\cite{astolfi1999}. We then compute the human actions using an attracting potential field at the goal position, and a repelling potential field centered at the robot. The algorithm then performs a set of optimizations, one for each explicit communication action. Taking advantage of modern multi-core CPUs, we can run these optimizations in parallel. Finally, the best explicit communication $u_c^*$ and robot movements (implicit communication) $\mathbf{u}_r^*$ are selected and executed.

We implemented the planning algorithm in C++, and used the software package NLopt~\cite{johnson2014} to perform numerical optimization. In our implementation, we chose a planning horizon of $N = 6$, and a time step of 0.5 seconds.

\section{Simulation Results} \label{sec: simulation}
In order to validate the proposed modeling and planning framework, we first collected human demonstrations and conducted a simulation experiment. In this section, we describe the specific scenario for our experiment, validation of the human model, and simulation results.

\subsection{Social Navigation Scenario}
We consider the scenario shown in Fig.~\ref{fig: intro} for our simulation and experimental evaluation: a human and a robot move in (approximately) orthogonal directions and encounter each other. In this scenario, they have to coordinate with each other and decide who should pass first to avoid collision. We chose this scenario over the face-to-face encounter scenario because in the face-to-face encounter, there is a relatively strong social norm that people pass each other on the right hand side. We prefer to start with a scenario with no such bias to allow us better focus on the effect of communication. The robot can explicitly communicate its intent: to yield priority to the human (human priority) or not (robot priority). To numerically evaluate the model and compute the optimal plan, we need to define features leading to a learned reward function, which can then be used to model humans' continuous actions Eq.~\eqref{eq: max ent} as well as behavior modes Eq.~\eqref{eq: implicit model} of the human.

We select features that are commonly used to characterize social navigation of pedestrians~\cite{hoogendoorn2003}:
\begin{description}[leftmargin=0cm]
\item [Velocity.] Pedestrians tend to maintain a constant desired velocity. We use a feature that sums up the squared velocity along the trajectory:
\begin{equation}
f_{velocity} = \sum_{t=1}^{N} \norm{\st{\dot{x}}{h}}^2
\end{equation}
Here, and in the following equations, $||\cdot||$ represents L2 norm.

\item [Acceleration.] To navigate efficiently, pedestrians usually avoid unnecessary accelerations:
\begin{equation}
f_{acceleration} = \sum_{t=1}^{N} \norm{\st{\ddot{x}}{h}}^2 = \sum_{t=1}^{N} \norm{\st{u}{h}}^2
\end{equation}

\item [Distance to goal.] Pedestrians typically try to get to the target position as close as possible given a time horizon:
\begin{equation}
f_{goal} = \norm{\st[N]{x}{h} - {x}_{\text{goal}}}
\end{equation}

\item [Avoiding static obstacles.] Pedestrians avoid static obstacles in the environment:
\begin{equation}
f_{obstacle} = \sum_{t=1}^{N} \exp \left( - \frac{\norm{\st{x}{h} - \st{o}{\text{closest}}}^2}{D_o^2} \right)
\end{equation}
where $o_{\text{closest},t}$ is the position of the closest obstacle to the human at time $t$, and $D_o$ is a scaling factor.

\item [Collision avoidance with the robot.] Pedestrians tend to avoid each other in navigation. We assume that they avoid the robot in a similar fashion:
\begin{equation}
f_{avoidance} = \sum_{t=1}^{N} \exp \left( - \frac{\norm{\st{x}{h} - \st{x}{r}}^2}{D_r^2} \right)
\end{equation}

\item [Avoiding the front side of the robot.] In addition to avoiding the robot, we observe that humans tend to not cut in front of the robot, especially when they think that the robot has priority. This behavior is captured by the feature:
\begin{equation}
f_{front} = \sum_{t=1}^{N} \exp \left( - (\st{x}{h} - \st{x}{f})^\intercal \mathbf{D}_f^{-1} (\st{x}{h} - \st{x}{f}) \right)
\end{equation}
where $\st{x}{f}$ is a position in front of the robot, and $\mathbf{D}_f$ scales the Gaussian and aligns it with the robot's orientation.
\end{description}

\begin{figure}[t]
	\centering
    \includegraphics[width=0.5\textwidth]{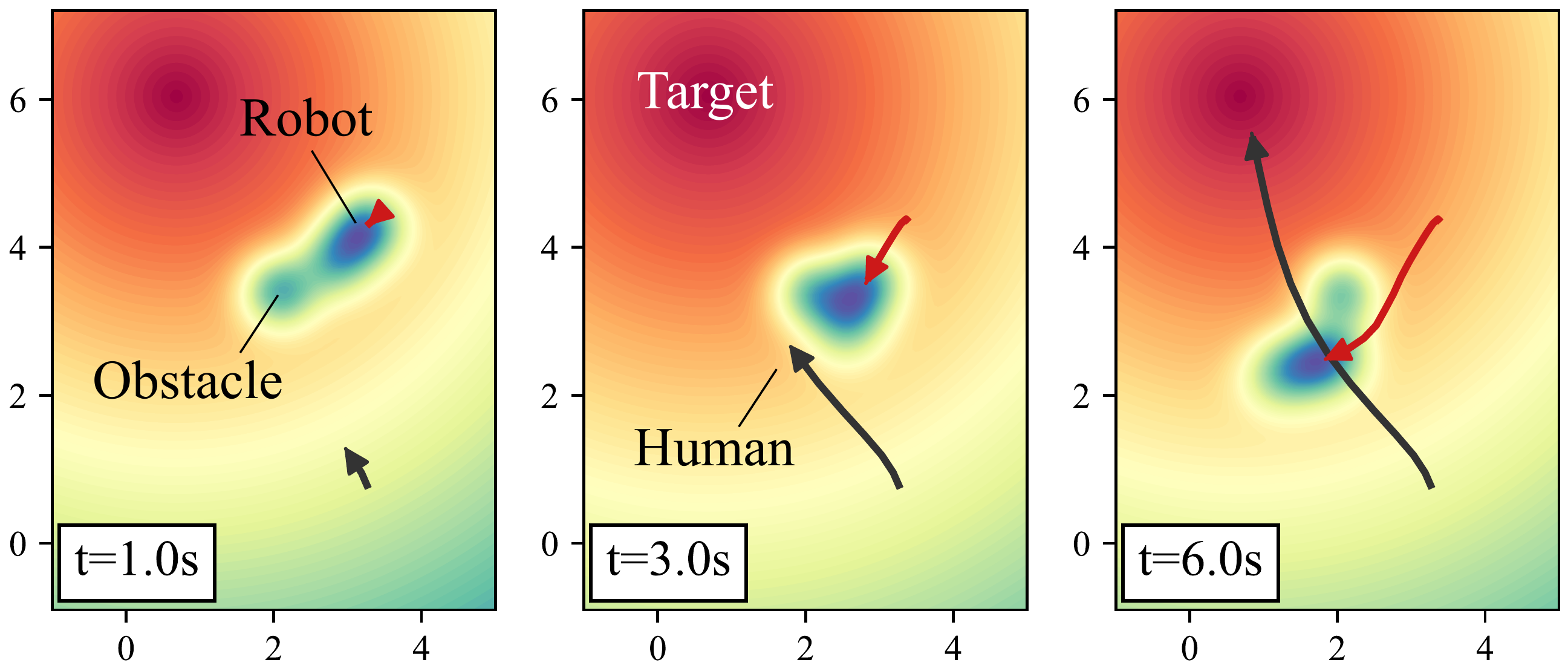}
    \caption{Visualization of (a subset of) the features. Warm color indicates high reward and cool color indicates low reward. Human trajectory is in black, and robot trajectory is in red. Arrows indicate the positions and moving directions at the specified time step.
    }
    \label{fig: features}
\end{figure}

The features (except for velocity and acceleration) are visualized in Fig.~\ref{fig: features}. Human and robot trajectories are from a demonstration we collected to train the IRL model. The figure shows that the human indeed avoided low reward regions (cool color) and navigated to the high reward region (warm color).

\begin{figure*}[t]
	\centering
    \includegraphics[width=0.95\textwidth]{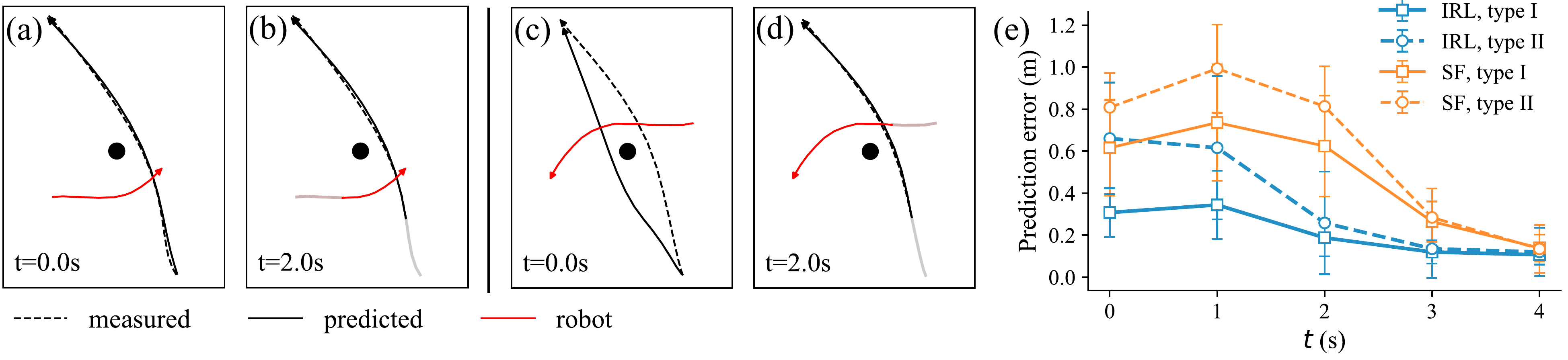}
    \caption{Sample predictions using the learned reward functions and cross validation. (a)-(b) show an example where the prediction matches the actual measurement. (c)-(d) show an example where initially the prediction doesn't match the measurement. We use the model to re-predict human actions at each time step, and the prediction starts to match measurement after two seconds. (e) Cross validation of prediction accuracy. Type I corresponds to cases where the model initially predicts the same homotopy class as the demonstration ((a), (b)), and type II corresponds to cases where the model initially predicts a different homotopy class ((c), (d)). Blue indicates errors of the IRL model, and orange indicates errors of the social force (SF) model. }
    \label{fig: cross validation}
\end{figure*}

\subsubsection*{Rewards for the Behavior Mode Model}
In general, it is challenging to to define the expected reward function of the human $R_h^\psi(\traj[-N:0]{r}, \traj[-N:0]{h})$, as it is not measurable. Here we design two reward functions based on our observations. Data driven methods like IRL can potentially produce more accurate results. Our goal, however, is to derive an approximate model that can inform the planning algorithm. Therefore, we prefer to start with simpler designs:

    \prg{Human Priority ($\psi = \psi_h$)} When the robot's intent is to yield priority to the human, it should avoid moving directly towards the human and getting too close. We use a reward function:
    \begin{equation}
        R_h^{\psi_h} = -\sum_{t = -N}^{0} \frac{\hat{n}_t \cdot (\st{\dot{x}}{r} - \st{\dot{x}}{h})}{\max(\norm{\st{x}{r} - \st{x}{h}}^2, 1.0)}
        \label{eq: implicit reward human priority}
    \end{equation}
    where $\hat{n}_t$ is a unit vector that points from the robot to the human: $\hat{n}_t = (\st{x}{r} - \st{x}{h}) / {\norm{\st{x}{r} - \st{x}{h}}}$. Note that here $t \leq 0$ means that the summation is over the previous time steps.
    
    \prg{Robot Priority ($\psi = \psi_r$)}  When the robot has priority, the human would expect it to move with a desired velocity, regardless of the state of the human:
    \begin{equation}
        R_h^{\psi_r} = -\sum_{t = -N}^{0} \frac{|\st{v}{r} - v_d|}{\max(\norm{\st{x}{r} - \st{x}{h}}^2, 1.0)}
        \label{eq: implicit reward robot priority}
    \end{equation}
    Here $\st{v}{r}$ is the robot's speed and $v_d$ is the desired speed. 
    
    

In Eqs.~(\ref{eq: implicit reward human priority}) and (\ref{eq: implicit reward robot priority}), the reward at each time step is divided by $\max(\norm{\st{x}{r} - \st{x}{h}}^2, 1.0)$ because when the robot is relatively far from the user, its actions carries less information about its intent. We use the $\max()$ operation to achieve numerically stable  results.

\subsection{Human Model Evaluation}
\prg{Data Collection for Learning Human Model}
We collected navigation demonstrations from four users. Each user was asked to walk back and forth between two target locations (setup is similar to that of Section~\ref{sec: experiment}). At the same time, a mobile robot (TurtleBot2, Open Source Robotics Foundation, Inc.) also moved between a set of target positions in the environment. The target positions were selected so that the robot would cross path with the user. The user was allowed to walk freely as in normal human environment, while avoiding collision with the robot.

We collected data for two scenarios: the human priority scenario and the robot priority scenario. When the user had priority, the robot would slow down and yield to the user upon encounter (we decreased the maximum allowable speed of the robot as it got close to the user). When the robot had priority, it would simply ignore the user. In each condition, the user was told the intent of the robot, so the user always knew the robot's true intent. To distract the user from focusing solely on the robot, we asked the user to remember two numbers at one target location, and answer arithmetic problems using the two numbers at the other target location.

For each user, we recorded 64 trials for each scenario (a trial is moving from one target location to the other). We used 80\% of the data to train the model, and the other 20\% as the testing set.

\prg{Prediction with Learned Reward Function}
Using MaxEnt IRL, we recover the human reward functions in human priority and robot priority scenarios. The reward functions can be used to compute the most likely continuous navigation actions using Eq.~\eqref{eq: most likely actions}. This computes the actions over a fixed time horizon. To predict the entire trajectory, we use an MPC approach: we compute the most likely actions starting from time step $k$, predict the human state at $k+1$, and recompute the most likely actions starting from $k+1$.

Fig.~\ref{fig: cross validation} presents two prediction examples. In the first example ((a)-(b)), the predicted trajectory matches the measured trajectory very well (same homotopy class, type I). In the second example ((c)-(d)), the model initially predicts a trajectory that passes the obstacle on a different side (different homotopy class, type II). Although the prediction does not match the measured trajectory, it is still reasonable. In this example, both the predicted and measured trajectories pass behind the robot. Passing on either side of the obstacle has little effect on the reward and is almost equally likely. When predicting at $t=2.0$ s, because the human already started to move toward one side of the obstacle, the model is able to 
generate a more accurate prediction.

The cross validation result is shown in Fig.~\ref{fig: cross validation}(e). The prediction error is calculated as the average Euclidean distance between the predicted and measured trajectories:
\begin{equation}
    e = \frac{1}{N'} \sum_{t=1}^{N'} \norm{\st{x}{\text{pred}} - \st{x}{\text{meas}}}
\end{equation}
where $N'$ is the total number of time steps. We compute the prediction error separately for cases where the initial prediction is in the same homotopy class as the measurement (type I), and cases where the initial prediction and the measurement are in different homotopy classes (type II). It can be observed that the initial prediction error is much smaller for type I, but both types becomes more accurate as the human gets closer to the goal. Compared with a social force model we used previously~\cite{Che2018}, the model described here performed better for this scenario.

\prg{Evaluation of the Behavior Mode Model}
\begin{figure*}[t]
	\centering
    \includegraphics[width=1.0\textwidth]{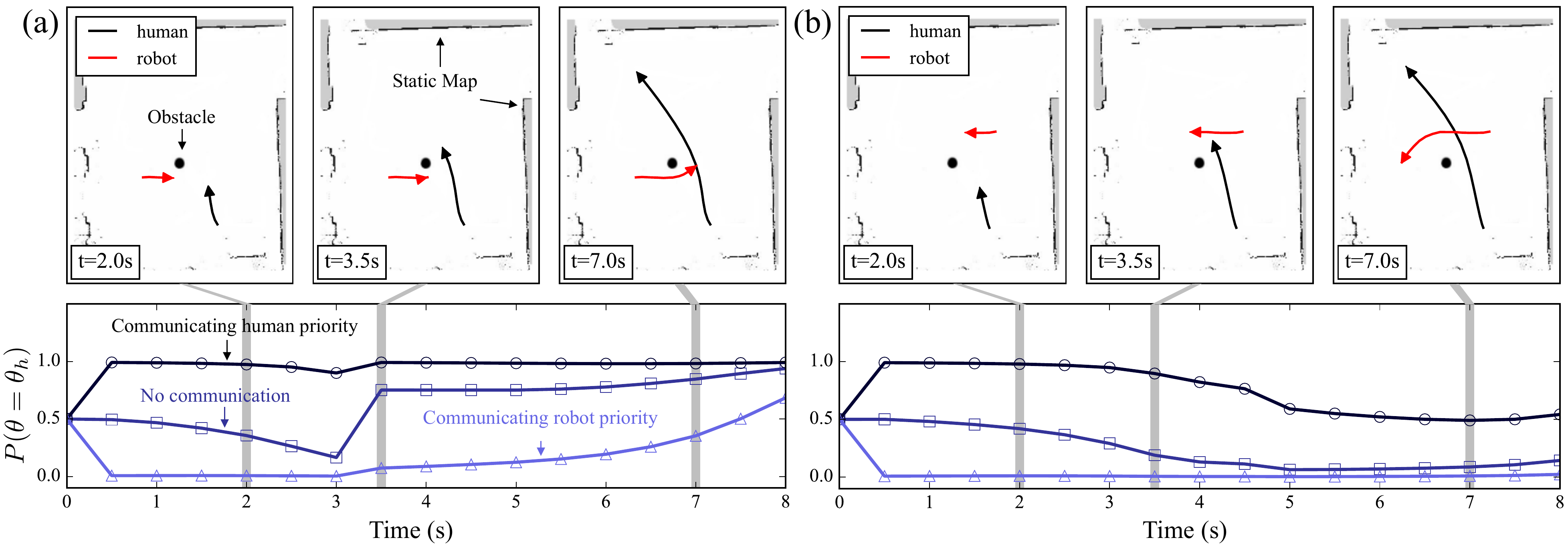}
    \caption{Demonstrations of the behavior mode model in different scenarios. (a) A scenario where the robot slowed down to let the user pass first. Top plots show the trajectories of the robot and the user, and a map of the environment at three different time steps. Bottom plot shows the user's belief over the robot's intent (human priority) over time, given different explicit communication at $t=0.5$ s, predicted by our model. (b) A scenario where the robot didn't slow down and passed first.}
    \label{fig: belief model examples}
\end{figure*}
In addition to continuous trajectories, our model can also predict behavior modes, or equivalently, the belief of the human over the robot's intent. As it is impossible to measure this belief in an experiment, we aim to show that the prediction is reasonable with a few test scenarios.

Fig.~\ref{fig: belief model examples}(a) presents a scenario where the robot slowed down to let the user pass first. The scenario is illustrated with the three plots in the top row, showing trajectories and the environment at different time steps. The data is from one demonstration we collected for training the IRL model. The plot in the bottom row shows the predicted belief $p(\theta=\theta_h)$ over time, given no explicit communication, and two different explicit communication actions at $t_c = 0.5$ s: communicating human priority ($u_c = \theta_h$) and communicating robot priority ($u_c = \theta_r$). In the case of no communication, the belief decreases initially, but rises up as the robot stops to yield to the user. This result matches the intuition that the user can infer the robot's intent (human priority) to some extent by observing its movement (slowing down). When the robot communicates human priority at the beginning, the belief jumps to a high value and maintains this value afterwards. An interesting test scenario is when the robot communicates robot priority, which is not its true intent. According to our model, initially the belief drops, indicating that the user believes the explicit communication. However, when the robot slows down to let the user pass first, the belief starts to increase, indicating that the user starts to believe otherwise as she observes the robot movements. 

We also tested scenarios where the robot's intent is to give itself priority. An example is shown in Fig.~\ref{fig: belief model examples}(b). Similarly, we show the change in belief given different explicit communication in the bottom plot. These examples demonstrate that our model can capture the effect of both explicit and implicit communication, and the prediction matches our intuition.

\subsection{Case Study in Simulation}
\begin{figure*}[t]
	\centering
    \includegraphics[width=1.0\textwidth]{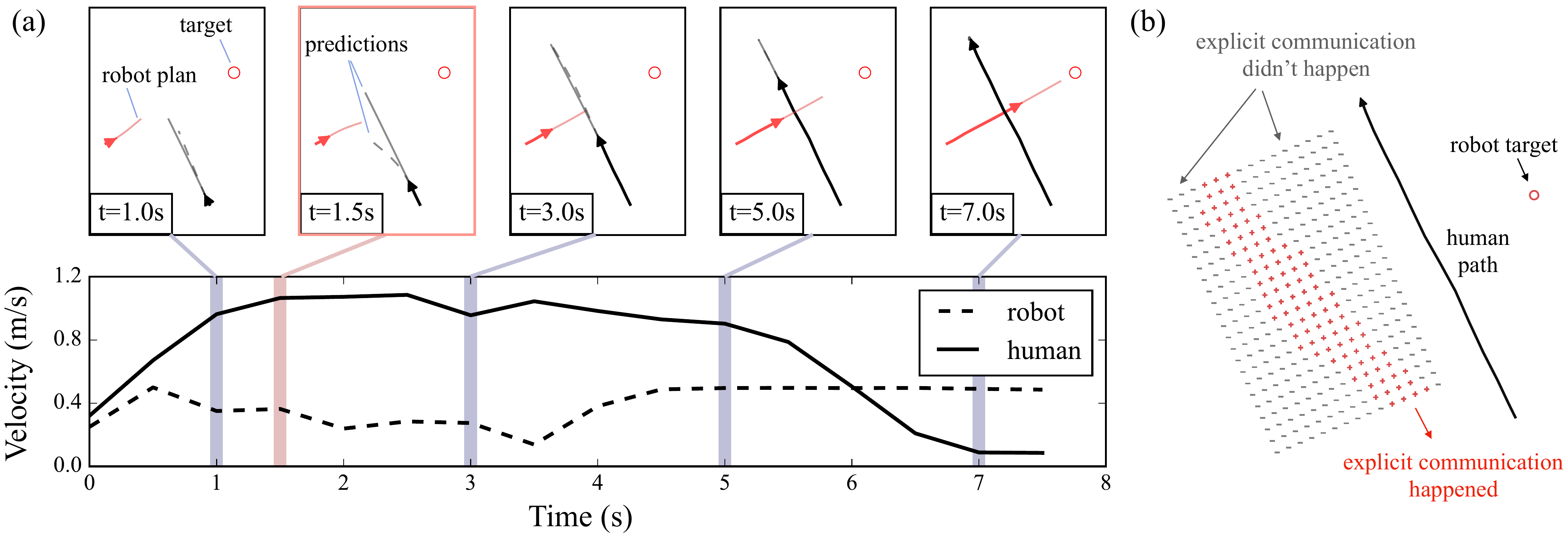}
    \caption{(a) Simulated scenario using the proposed interactive planner. In this scenario, the robot slows down and explicitly communicates its intent (human priority) to the human. Top: trajectory snippets at five individual time steps. Black and red arrows represent the positions and moving directions of the human and the robot at the given time step. Thin, solid red lines represent the robot's planned trajectories, thin solid/dashed black lines represent predicted human trajectories. Bottom: Velocity profile of the robot and the human. (b) Various starting positions of the robot vs. whether explicit communication is used by the planner for the same scenario. Red region represent the starting positions where explicit communication is used.}
    \label{fig: simulation example}
\end{figure*}

Before conducting a user study with a physical mobile robot (described Section~\ref{sec: experiment}), we first tested our algorithm in simulation. The purpose of the simulation is two-fold: first to validate that the proposed algorithm can work in idea setups, 
and second to select appropriate parameters that produce reasonable behaviors.

Fig.~\ref{fig: simulation example}(a) illustrates a simulated scenario where the human and the robot move along orthogonal paths. Here the simulated human user follows a predefined trajectory, and the trajectory is deterministic regardless of the robot's actions. While this is not realistic, the purpose is to test whether the planner can generate reasonable robot behaviors. We will further validate the effectiveness of the plan with real-world user studies. The subplots in the top row shows the scenario at 5 different time steps, and the subplot in the bottom row shows the speed of the user and the robot over time. We can observe that the robot starts to slow down at $t = 1$ s, and explicitly communicates its intent to the user at $t = 1.5$ s. The robot continues moving slowly to allow the user pass first, and then speeds up towards its goal.

\begin{figure*}[thpb]
	\centering
    \includegraphics[width=\textwidth]{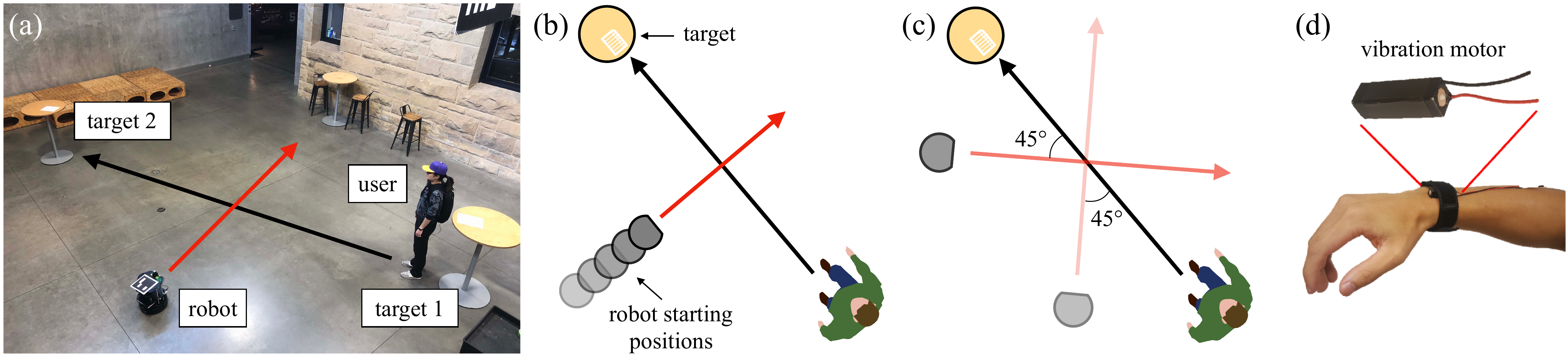}
    \caption{(a) Experimental field. (b-c) Different starting and goal positions for the robot. (d) The wearable haptic interface and the vibration motor.}
    \label{fig: setup}
\end{figure*}

To understand why the planner does this, we visualize the generated plan and the predictions of human movements, as shown in Fig.~\ref{fig: simulation example}(a). As the robot speeds up and approaches the intersection, the user becomes unsure of how to avoid the robot. This is reflected in the second subplot in the first row, where the user's possible future trajectories diverge. In this simulation, we set the reward function coefficients $c_h \gg c_r$, so that the planner cares more about the human user's efficiency and comfort. As a result, the planner explicitly communicates human priority to the user to minimize the chance that the user would slow down to yield to the robot.

We tested the planner with various robot starting positions and target positions. Fig.~\ref{fig: simulation example}(b) presents the relationship of robot starting position and whether explicit communication is generated by the planner, for one specific target position. We can see that explicit communication is only used if the robot starts within the banded region in red. When the robot starts too close to the intersection (upper-right grey region), the planner predicts that the user will let the robot pass first as it is closer to the intersection. When the robot starts too far (bottom-left grey region), the planner predicts that the user will pass first. Explicit communication is used only when it becomes ambiguous who should pass first.

We also studied the effect of the coefficients $c_h$ and $c_r$ in the reward function in Eq.~(\ref{eq: robot reward function}). Setting $c_h \gg c_r$ resulted in a submissive robot that would yield to the user when there was a potential collision. Conversely, setting $c_h \ll c_r$ resulted in aggressive robot behaviors. When $c_h \approx c_r$, the robot tends to be submissive. In our implementation, the robot is usually slower than the human, so it is often more efficient to let the user pass first.
Based on the simulation results, we decided to experimentally test two sets of parameters: $c_h : c_r = 1:5$ as \emph{robot-prioritized} scenario, and $c_h : c_r = 2:1$ as \emph{human-prioritized} scenario.

\section{Experimental Results} \label{sec: experiment}
Simulation results from the previous section have demonstrated that the planning framework can generate appropriate implicit and explicit communication actions. This section presents a real-world user study to further test the effectiveness of the proposed planning framework.

\subsection{Experimental Setup}
We again used a TurtleBot2 as the mobile robot platform. The robot was equipped with a Hokuyo URG-04LX-UG01 laser range finder, and an ASUS Zenbook UX303UB as the onboard computer. The planning algorithm is implemented on a desktop computer with Intel Core i7 processor and 16GB RAM. The two computers communicate with each other through a wireless network.

The onboard computer processes the laser range finder readings to localize the robot and estimate the position and velocity of nearby pedestrians. We localize the robot in a static map using laser-based Monte Carlo Localization~\cite{thrun2001}. The robot detects and tracks pedestrians using a leg detection algorithm provided by ROS~\cite{quigley2009}. The algorithm first attempts to segment leg objects from laser range finder readings with a pre-trained classifier, then pairs nearby legs and associates the detection with a tracker for each person. The localization and tracking results are sent to the desktop computer that runs our planning algorithm. The planning algorithm computes the plan and sends it back to the robot.

The explicit communication is displayed to the user via a wearable haptic interface that consists of a single vibration motor (Haptuator Mark II, shown in Fig.~\ref{fig: setup} (d)). The interface is capable of rendering distinct signals by modulating the vibration pattern. In this experiment, we display haptic cues with different vibration amplitudes and durations to indicate the robot's intent. Robot priority is represented by a single long vibration (duration 1.5s, max current 250 mA), and human priority is represented by 3 short pulses (0.2 sec vibration with 0.2 sec pause in between, max current 150 mA). To train the user interpreting the haptic cues, we first display cues in random order (3 trials for each vibration pattern) and explain the meaning. Then we test the user with randomized cues. We found that all users can recognize each cue correctly. The training process is done for each user before the actual experiment.

The experiment field is illustrated in Fig.~\ref{fig: setup}(a), which is a room of size 8 m$\times$10 m. Two tables are placed at the two ends of the field as the targets for the user. To distract the user from focusing solely on the robot, we place questionnaires on the tables, and ask the user to remember and answer arithmetic questions. The entire experiment is recorded with an overhead camera (GoPro Hero 4, recording at 60 Hz), and we post-process the video to extract the trajectories of the user and the robot. To facilitate tracking, we ask the user to wear a purple hat, and we attach an ArUco marker~\cite{Aruco2014} on the top of the robot.

\subsection{Experimental Design}

\begin{figure*}[t]
	\centering
    \includegraphics[width=\textwidth]{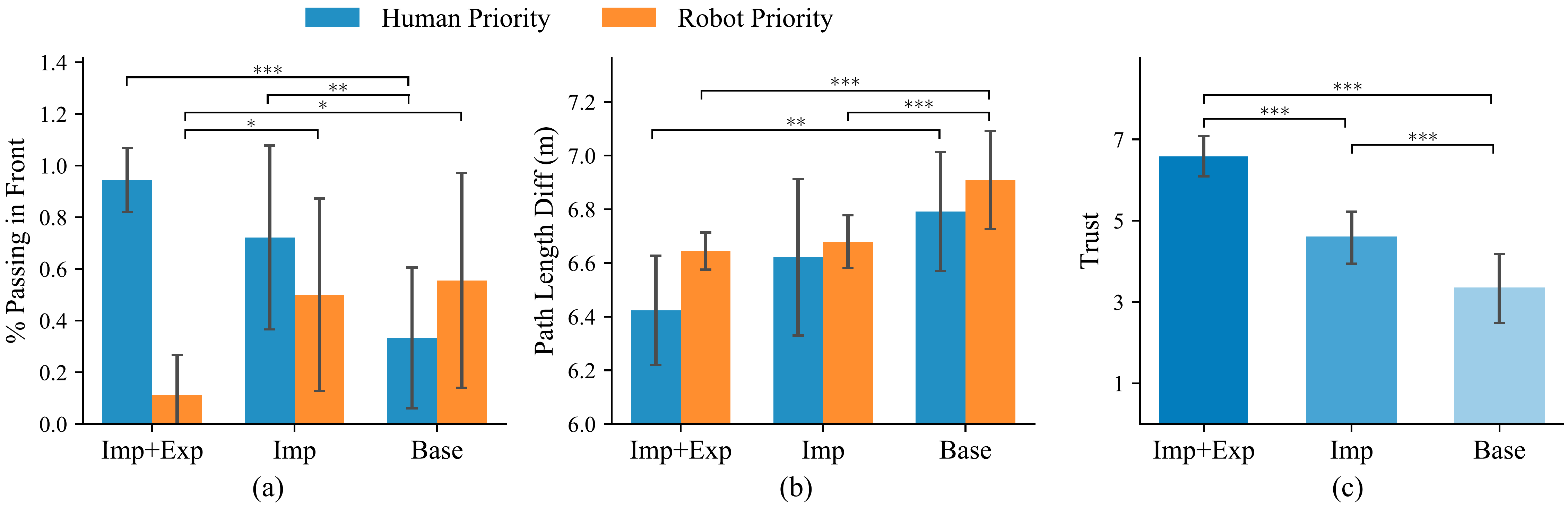}
    \caption{Comparison of different metrics for three experimental conditions. (a) Percentage of time that the user passed in front of the robot. (b) Average path length. A base length (6.2 m) is subtracted from all for visualization purpose. (c) The user's trust in the robot. Brackets indicate statistical significance (*$p < 0.05$, **$p < 0.01$, ***$p < 0.001$). }
    \label{fig: stats}
\end{figure*}

We conducted a $3 \times 1$ within-subject experiment, manipulating the communication mode (baseline, implicit only, explicit + implicit) and measuring users' path lengths, whether they pass the robot in the front, and their trust in the robot. Details of the experiment design and procedure are described below.

\prg{Manipulated Factors} We manipulate one factor: the \textbf{\textit{communication mode}}. The user's performance under each communication mode reflects the effectiveness of the communication method. In addition, the experimental trials consist of different \textit{task priorities} and \textit{robot starting positions}. We randomize the order of task priority so that the user does not know the robot's true intent beforehand. We also vary the starting position of the robot to create richer interaction scenarios and prevent the user from developing a fixed pattern (e.g. always pass in front of the robot regardless of communication). Overall, the experiment is divided into three sessions based on the communication mode: 
\begin{itemize}
    \item \textbf{\textit{explicit + implicit}}: the robot communicates its intent to the user both explicitly (via haptic feedback) and implicitly (by changing speed and direction), using a model to predict the user's movement.
    \item \textbf{\textit{implicit only}}: the robot does not plan for explicit communication, but still changes speed and direction according to a model that predicts the user's movement.
    \item \textbf{\textit{baseline}}: the robot simply performs collision avoidance with the user without predicting the user's movement.
\end{itemize}

Each session consists of 20 trials. In each trial, the human user and the robot both move from a target position to a goal position, as shown in Fig.~\ref{fig: setup}(b). The robot is assigned one of the two task priorities in each trial: robot priority ($c_h : c_r = 1:5$) or human priority ($c_h : c_r = 2:1$). We also vary the starting position of the robot. The starting positions can be classified as \textbf{\textit{far}} (4 out of 20 trials), \textbf{\textit{close}} (4/20) and \textbf{\textit{normal}} (12/20), based on the distance to the user's starting position (Fig.~\ref{fig: setup}(b)). There are equal numbers of high-priority trials and low-priority trials for each type of starting position. The order of the trials for each communication mode is pseudo-randomized. We counter-balance the order of three communication modes among all users using a Latin square design. The first 10 trials are treated as training trials and not included for analysis. In the training trials, we will reveal the robot's true intent to the user after each trial. We never tell the robot's intent to the user in the experimental trials. The purpose of the training trials is to familiarize the user with the experimental procedure and the robot's behavior. After the experiment, we ask the user to rate her trust in the robot on a scale from 1 to 7, 1 being the least trust and 7 being the most trust. The exact question we use in the survey is ``Please rate your overall trust in the robot in the first/second/third condition''. There are 7 choices under each question and we explain to the user the meaning of the scale.

In addition to the orthogonal encounter scenario, we also tested the planner in other experimental setups. Fig.~\ref{fig: setup} (c) presents human-robot encounter at different angles ($+45^\circ$ and $-45^\circ$). We found that the planner worked similarly in these scenarios.

\prg{Dependent Measures} We measure the user's path length for each trial. Path length is commonly used to estimate the user's efficiency and effort in navigation tasks. We also measure whether the user passes the robot in the front, which is an indicator of whether the user understands the robot's intent correctly (the user is expected to pass the robot in the front in human priority trials and from the behind in robot priority trials, if they understand the robot's intent correctly. This measurement reflects the transparency of the robot's behavior. Finally, we measure the user's trust in the robot. Trust is measured using a post-experiment survey as described previously. In the session where explicit communication is allowed, we also record whether explicit communication happens and the time at which it happens.

\prg{Hypothesis} We hypothesize that:
\begin{enumerate}[label=\Roman*., widest=IV, align=left, leftmargin=*]
\item Using \textbf{\textit{explicit + implicit}} communication conveys the robot's intent more clearly than \textbf{\textit{implicit only}} and \textbf{\textit{baseline}}, such that users will elect to pass in front of or behind the robot as appropriate for a given priority.
\item The user's average path length is shorter when the robot plans for communication with the human model (\textbf{\textit{explicit + implicit}} and \textbf{\textit{implicit only}} modes).
\item The user's overall trust rating in the robot is higher when the robot plans for \textbf{\textit{explicit + implicit}} communication, compare to the other two conditions.
\end{enumerate}


\prg{Subject Allocation} A total of 12 people (7 males and 5 females, age from 23 to 33) participated in the experiment after giving informed consent, under a protocol that was approved by the Stanford University Institutional Review Board. We used a within-subjects design and counterbalanced the order of the three sessions.

\subsection{Analysis and Results}
\begin{table}[t]
\caption{Mean and standard deviation of measurements. For the first two measurements, the first row is Human Priority and the second row is Robot Priority.}
\label{table: exp measurements}
\centering
\begin{tabular}{|@{\hspace{3pt}}l@{\hspace{3pt}}|c|c|c|}
\hline
Measure & Imp+Exp & Imp & Base\\
\hline\hline
\multirow{2}{1.5cm}{\% Passing in Front} & $0.94\pm0.12$ & $0.72\pm0.36$ & $0.33\pm0.27$\\ \cline{2-4}
& $0.11\pm0.15$ & $0.50\pm0.37$ & $0.55\pm0.41$\\
\hline
\multirow{2}{1.5cm}{Path Length} & $6.42\pm0.20$ & $6.62\pm0.29$ & $6.79\pm0.22$\\ \cline{2-4}
& $6.64\pm0.07$ & $6.68\pm0.10$ & $6.91\pm0.18$\\
\hline 
Trust  & $6.58\pm0.49$ & $4.58\pm0.64$ & $3.33\pm0.85$\\
\hline
\end{tabular}
\end{table}

Fig.~\ref{fig: stats} and Table~I summarizes major results. We describe the analysis and results in detail in the following paragraphs.

\prg{Understanding Robot Intent} To characterize how often the user correctly understood the robot's intent, we compute the percentage of trials that the user passed in front of the robot for each communication mode and task priority. A one-way repeated measures ANOVA revealed a significant effect on percentage passing in the front for the communication mode ($F(2, 33) = 14.61, p < .001$ for human priority trials, $F(2, 33) = 5.75, p = .007$ for robot priority trials). We performed the statistical test separately for robot and human priority trials. We performed a post-hoc analysis with Tukey HSD to determine pairwise differences. For human priority trials, results show that the baseline mode is significantly different from the implicit only mode ($p = .005$) and the implicit + explicit mode ($p < .001$). For robot priority trials, the implicit + explicit mode is significantly different from the implicit only mode ($p = 0.027$) and the baseline mode ($p = 0.01$). This supports Hypothesis I that the user can better understand the robot's intent with explicit communication.

\prg{Path Length} Path length is a commonly used metric to measure efficiency. We computed average path length (over all trials of each condition and robot intent) of each user. Similar to the metric of passing in the front, we performed a one-way repeated measure ANOVA analysis, and results show a significant effect ($F(2, 33) = 6.36, p = .005$ for human priority trials, $F(2, 33) = 14.18, p < .001$ for robot priority trials). Post-hoc comparisons reveal that, for human priority trials, the average path length of the implicit + explicit mode is significantly shorter than the baseline mode ($p = .003$). For robot priority trials, the average path length of the baseline mode is significantly longer than the implicit only mode ($p < .001$) and implicit + explicit mode ($p < .001$). The result suggests that the user can navigate more efficiently when the robot plans for communication (Hypothesis II).

\prg{Trust} Past research has shown that human would trust the robot more if it can express its intent via motion~\cite{dragan2015} or other forms of communication~\cite{baraka2018}, and we expect similar results in our experiment. Here users' trust in the robot was measured with a post-experiment survey, as explained in the previous subsection. A one-way repeated measure ANOVA analysis revealed a significant effect for communication mode ($F(2, 33) = 64.5, p < .001$). Post-hoc Tukey HSD analysis showed that all three pairs are significantly different from each other ($p < .001$ for all), which supports Hypothesis III.



\begin{figure}[t]
	\centering
    \includegraphics[width=0.4\textwidth]{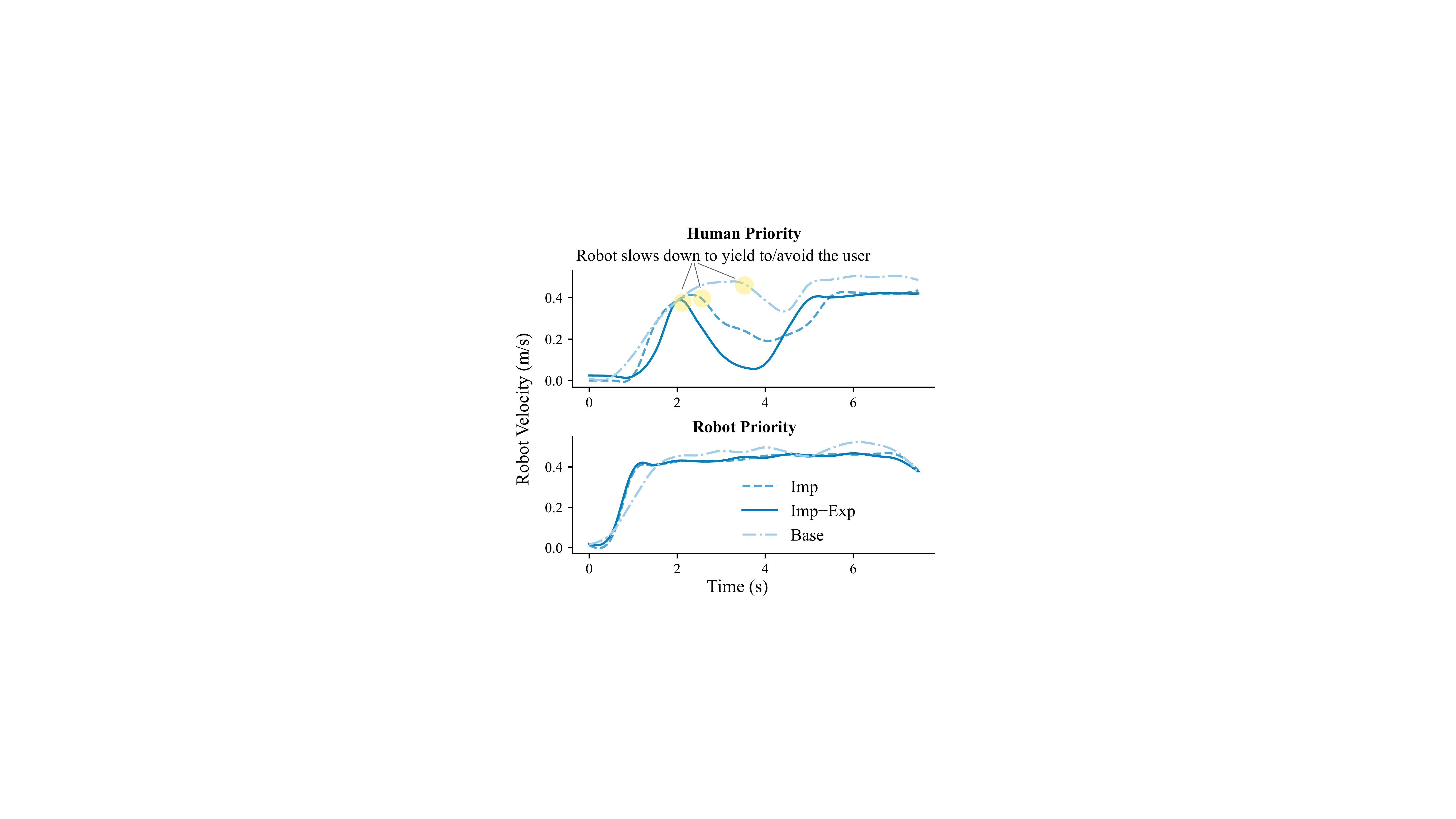}
    \caption{Velocity profiles from sample trials. Top: human priority trials. Bottom: robot priority trials.}
    \label{fig: velocity profile}
\end{figure}

\subsection{Effect of Communication}
Fig.~\ref{fig: stats}(a) shows that when there was both implicit and explicit communication, users were able to understand the robot's intent most of the time and acted accordingly -- users passed in front of the robot 92\% of the time in human priority trials, and only 9\% of the time in robot priority trials. When there was only implicit communication, users were more confused, especially in robot priority trials (about 50\% of the time users passed in front of the robot despite that the robot's intent was not to yield). When the robot performed collision avoidance naively (baseline condition), users were the most confused, and acted more conservatively.

Fig.~\ref{fig: velocity profile} depicts the robot's velocity over time from sample trials of different communication modes and task priorities. In human priority trials (top row), when the robot planned for communication (implicit only and implicit + explicit modes), it started to slow down relatively early, compared to the baseline mode. By doing so, the robot implicitly expressed its intent. Indeed, significantly more users chose to pass in front of the robot with these communication modes (Fig.~\ref{fig: stats}(a)). The reason that the planner generated this behavior is that, with the human model, the planner was able to predict the effect of robot motions on the user's navigation behavior. More specifically, our human model would predict that, given the robot slows down early, the human is more likely to believe that the robot is yielding priority and will choose to pass first, which would generate higher expected rewards. This encouraged the robot to slow down earlier. On the other hand, the baseline planner did not perform any prediction and simply solved for a collision-free trajectory. As a result, the robot did not slow down until it got very close to the user, which caused the user to believe that it did not want to yield. The result suggests that legible motions emerge when reactions of the human are taken into consideration during planning.

Our results also suggest that users can navigate more efficiently when the robot planned for communication. Fig.~\ref{fig: stats}(b) shows that for human priority trials, users' path length is significantly shorter in the implicit + explicit mode than the baseline mode. For robot priority trials, both the implicit + explicit mode and the implicit only mode result in significantly shorter path lengths than the baseline mode. Knowing the robot's intent, users can better coordinate their movements with the robot to navigate more efficiently. In the implicit + explicit mode, users in general got feedback early in the trial. In the implicit only mode, users had to infer the robot's intent, and tended to be less efficient. This result does not show statistical significance; users' behaviors have high variance in the implicit only mode. In Fig.~\ref{fig: stats}(a), we observe the same high variance in the implicit only mode.

An important feature of the study is that we never explained to the user how the robot may move differently in three communication modes. One of the main purpose of the training trials is to let the user figure out the robot's movement pattern. We observe that users in general were able to adapt to the robot's communication mode in a few trials. Most users started conservatively, trying to keep relatively large distances from the robot. In the implicit + explicit mode, most users adapted to trust the explicit communication quickly. In the baseline mode, users made their decisions largely based on the initial distance from the robot, as they found it difficult to tell the robot's intent from its motion. Interestingly, in the implicit only mode, we observed that users' behaviors varied based on their strategy of interacting with the robot. Aggressive users tend to realize earlier that the robot would slow down when it wanted to yield priority; they were more likely to cut in front of the robot. This helped them to interpret the implicit communication. Timid users, on the other hand, tend to keep larger distances or let the robot pass first. Sometimes this behavior resulted in the robot not slowing down in human priority trials. Unlike legibility planning as in~\cite{dragan2013legibility}, our algorithm does not explicitly optimize legibility. In addition, the algorithm responds to human actions in real time. As a result, the robot would not wait if the user is too far, even in human priority trials, which could confuse the user. This explains in part the high variance observed in the implicit only mode.

Note that in the implicit + explicit mode, the robot may not always use explicit communication. Again, this is the result of optimizing the overall reward function. We assigned a negative reward for performing explicit communication to prevent over-communicating. In trials where the robot started too close to or too far  from the intersection point, it often chose to not use explicit communication because the chance of collision if extremely low.

Finally, we show that users gained more trust in the robot with the proposed planner in Fig.~\ref{fig: stats} (c). We found statistical significance between each pair of communication modes. When the robot communicated both implicitly and explicitly, users trusted the robot the most. This is also the condition where users understood the robot's intent the best (based on the previous analysis shown in Fig.~\ref{fig: stats}(a)). The correlation suggests that the easier users can understand the robot's intent, the more they trust the robot -- transparency is beneficial in social navigation. When users do not trust the robot, they tend to act conservatively and be less efficient (fewer trials where the user passed in front of the robot, and longer path lengths in the baseline mode). Note that the trust measurement in this work is exploratory, as we only asked users to rate their overall trust over the robot. In order to obtain more accurate results, a more specific and validated questionnaire should be used.

\section{Conclusion and Future Work} \label{sec: discussion}
In this paper, we presented a planning framework that leverages implicit and explicit communication to generate efficient and transparent social navigation behaviors. This approach for mobile robot proactive communication is inspired by humans' ability to use both explicit and implicit communication to avoid collisions during navigation. The planner relies on a new probabilistic model that predicts human movements. We evaluated the planner both in simulation and in a user study with a physical mobile robot. Results showed that the proposed planning algorithm can generate proactive communicative actions, which better expressed the robot's intent, reduced users' effort, and increased users' trust of the robot compared to collision avoidance with and without a model that predicts users' movements.

There are numerous ways to expand on this work. First, the model of human navigation behavior makes certain assumptions and approximations as described in Section~\ref{sec: human model}. As a result, the planner can not deal with certain scenarios (e.g., if the human suddenly stops and remains stationary). Expanding the data collection for IRL, including additional behaviors, and relaxing assumptions, would improve the model and thus the planner. Second, our planner is computationally expensive. One approach to reducing the computational complexity of planning is to use sampling-based methods, instead of performing optimization. Third, we can generalize our approach to consider richer interaction scenarios and other communication modalities. We are also interested in identifying the advantages of combining implicit and explicit communication in other application scenarios. For example, in collaborative tasks such as assembly, prompt explicit communication can be used to align each agent's goal, while implicit communication such as forces can improve physical interactions between partners. Finally, we would like to explore in future work how implicit and explicit communication affects users' trust towards the robot. The result in this work is an initial demonstration of the benefit of communication in terms of trust. However, more studies, such as looking into specific aspects of trust~\cite{hancock2011}, are needed to gain deeper understanding.

While this work only explored a relatively simple scenario, the result has clearly demonstrated the benefit of combining implicit and explicit communication in robot social navigation. We believe that the computational framework for modeling and planning communicative actions can potentially be adapted to other applications and improve interactions between robots and humans.


\ifCLASSOPTIONcaptionsoff
  \newpage
\fi

\bibliographystyle{IEEEtran}
\bibliography{IEEEabrv,bibliography}

\end{document}